\documentclass[letterpaper, 10 pt, conference]{ieeeconf}  

\IEEEoverridecommandlockouts                              

\usepackage{subfiles}
\usepackage{moresize}
\usepackage{amsmath}
\usepackage{adjustbox}
\usepackage{tabularx}
\usepackage{nomencl}
\usepackage{ragged2e}
\usepackage{amssymb}
\usepackage{mathtools}
\usepackage{graphicx}
\usepackage{float}
\usepackage{subcaption}
\usepackage{epstopdf}
\usepackage[tableposition=top]{caption}
\usepackage{array}
\usepackage{url}
\usepackage{algorithm}
\usepackage{algpseudocode}
\usepackage{diagbox}
\usepackage{multirow}
\usepackage[table]{xcolor}
\usepackage[framemethod=TikZ]{mdframed}
\usepackage[procnames]{listings}
\usepackage[hidelinks]{hyperref}
\usepackage{tcolorbox}
\mdfdefinestyle{frameStyle}{%
    roundcorner=5pt,
    backgroundcolor=white}

\algtext*{EndIf}
\algtext*{EndFor}
\algtext*{EndWhile}
\algtext*{EndFunction}

\newcommand{\assign}[0]{$\leftarrow$ }

\title{\LARGE \bf
Adaptive Compliant Robot Control with Failure Recovery for Object Press-Fitting
}

\author{Ekansh Sharma$^{\dagger}$, Christoph Henke$^{\ddagger}$, Alex Mitrevski$^{\dagger}$, and Paul G. Pl{\"o}ger$^{\dagger}$
\thanks{$^{*}$This work was conducted at the Institute for Business Cybernetics e.V as part of the MASON project (\url{https://mason-projekt.de/}), which is supported by the Federal Ministry for Economic Affairs and Climate Action under the funding IGF-Project no. 22403. The work was also supported by the b-it foundation.} %
\thanks{$^{\dagger}$Autonomous Systems Group, Department of Computer Science, Hochschule Bonn-Rhein-Sieg, Sankt Augustin, Germany
        {\tt\scriptsize ekansh.sharma@smail.inf.h-brs.de}, {\tt\scriptsize <aleksandar.mitrevski, paul.ploeger>@h-brs.de}}%
\thanks{$^{\ddagger}$Institute for Business Cybernetics e.V. at the RWTH Aachen University, Aachen, Germany
        {\tt\scriptsize christoph.henke@ifu.rwth-aachen.de}}%
}

\begin{document}

\maketitle
\thispagestyle{empty}
\pagestyle{empty}


\begin{abstract}
Loading of shipping containers for dairy products often includes a press-fit task, which involves manually stacking milk cartons in a container without using pallets or packaging. Automating this task with a mobile manipulator can reduce worker strain, and also enhance the efficiency and safety of the container loading process. This paper proposes an approach called Adaptive Compliant Control with Integrated Failure Recovery (ACCIFR), which enables a mobile manipulator to reliably perform the press-fit task. We base the approach on a demonstration learning-based compliant control framework, such that we integrate a monitoring and failure recovery mechanism for successful task execution.
Concretely, we monitor the execution through distance and force feedback, detect collisions while the robot is performing the press-fit task, and use wrench measurements to classify the direction of collision; this information informs the subsequent recovery process. We evaluate the method on a miniature container setup, considering variations in the (i) starting position of the end effector, (ii) goal configuration, and (iii) object grasping position. The results demonstrate that the proposed approach outperforms the baseline demonstration-based learning framework regarding adaptability to environmental variations and the ability to recover from collision failures, making it a promising solution for practical press-fit applications.
\end{abstract}


    \section{INTRODUCTION}
    \label{sec:introduction}

\begin{figure}[t]
    \begin{subfigure}{\linewidth}
        \centering
        \includegraphics[width=0.9\linewidth]{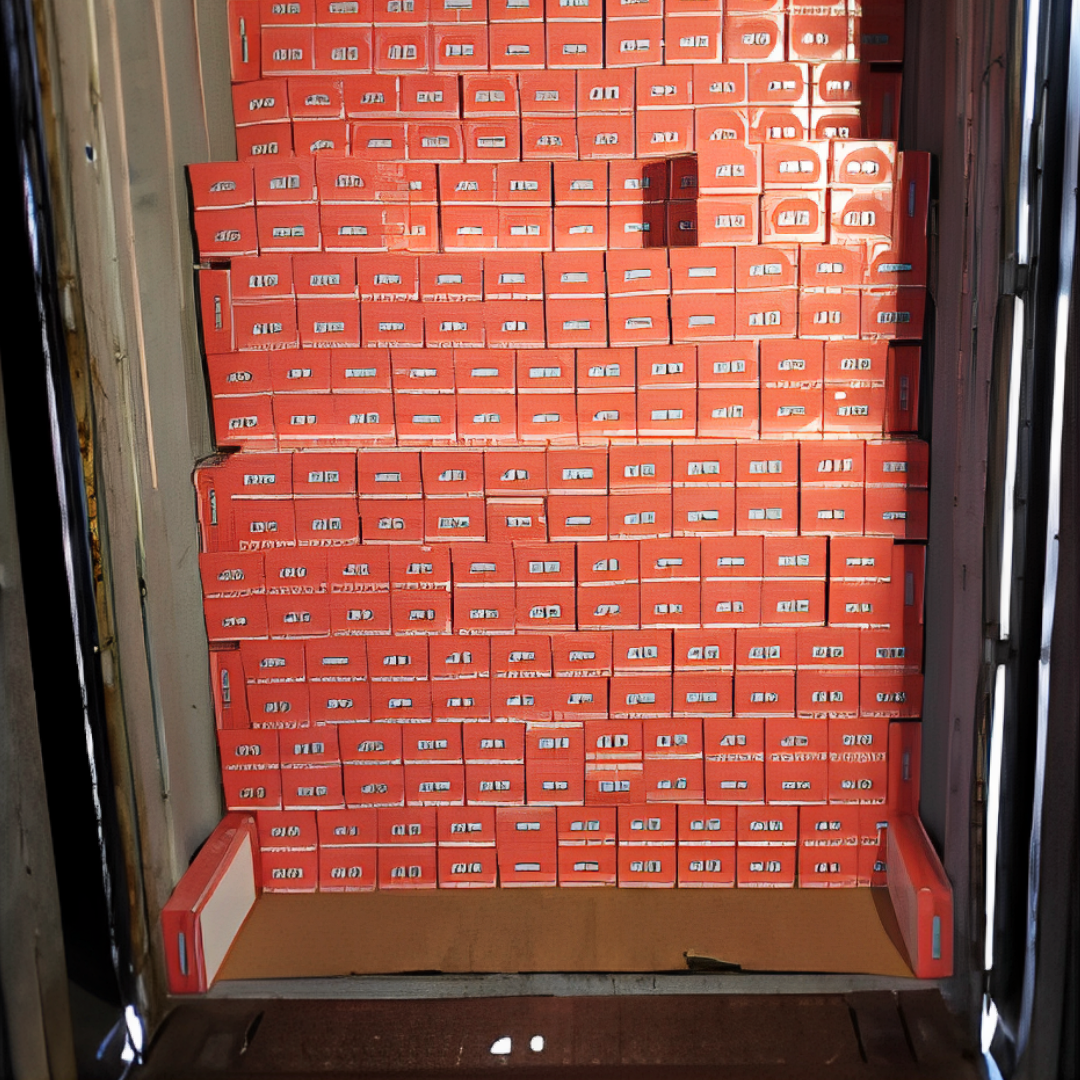}
        \caption{Container filled with milk cartons through manual press-fitting}
        \label{fig:shipping_container}
    \end{subfigure}
    \hfill
    \begin{subfigure}{\linewidth}
        \centering
        \includegraphics[width=0.9\linewidth]{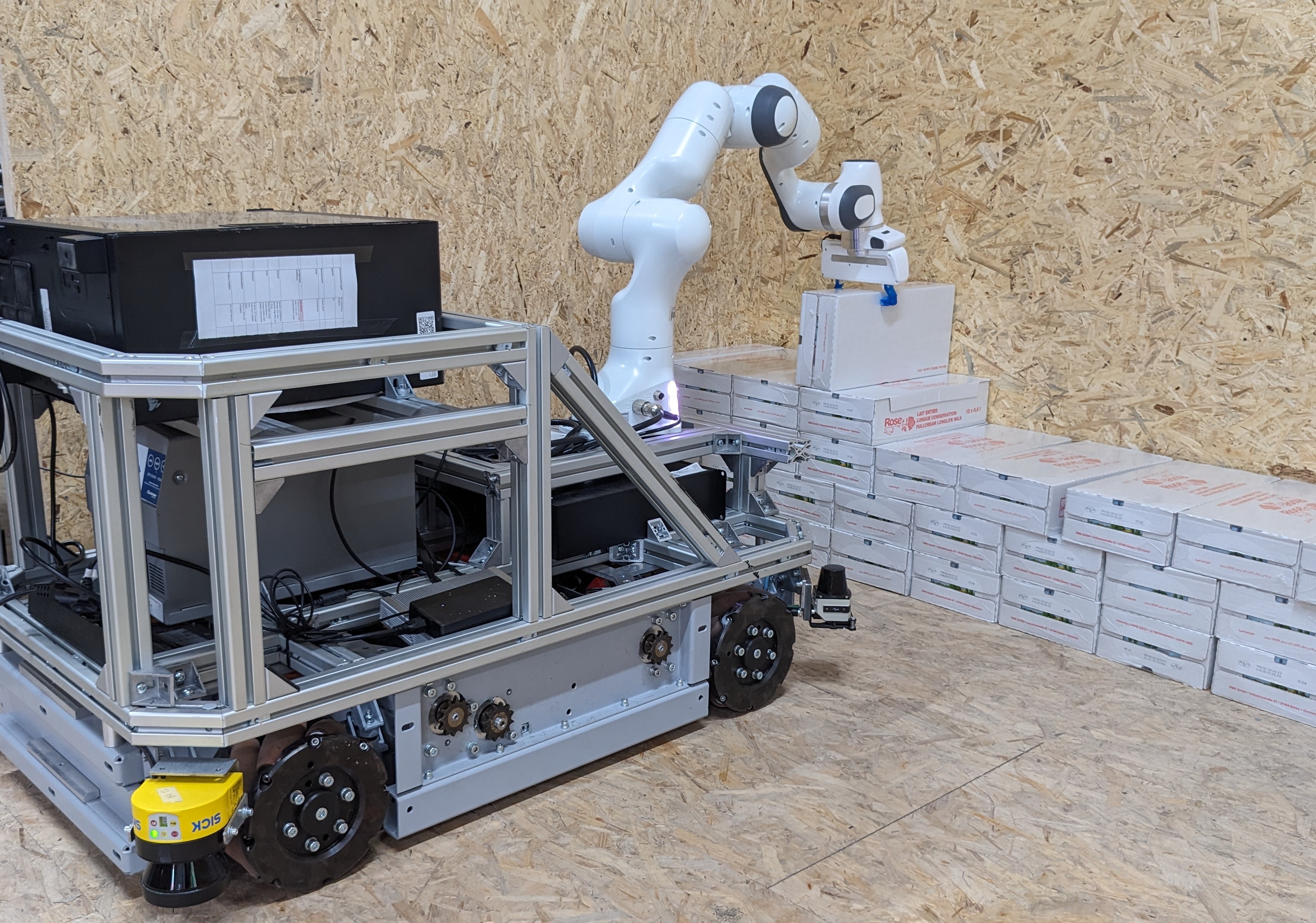}
        \caption{Custom mobile manipulator developed at the Institute for Business Cybernetics for automating the process}
        \label{fig:mobile-robot}
    \end{subfigure}
    \caption{Shipping container loading process}
    \label{fig:container-loading-process}
    \end{figure}

    The process of loading shipping containers with general cargo typically involves a press-fitting task, wherein the products are tightly packed with minimal clearance, as depicted in Fig. \ref{fig:shipping_container}.
    Press-fitting eliminates the need for additional packaging, such as pallets, which can be costly and take up valuable space within shipping containers. For dairy products, the task is usually performed manually in artificially cooled environments to ensure product quality.
    This manual process can, however, be physically demanding for workers, who must repetitively stack and push milk cartons. Furthermore, a shortage of skilled labor for this type of work is a persistent challenge in the industry \cite{ahn2022}. Automating the task using a mobile manipulator, as depicted in Fig. \ref{fig:mobile-robot}, can offer significant benefits in terms of efficiency and productivity; in particular, robots can work continuously without fatigue in temperature-controlled environments, resulting in a more reliable loading process.

    Performing the task with a mobile manipulator for loading shipping containers can be challenging due to the need for contact-rich object manipulation and susceptibility to environmental variations and collisions. Existing research on the peg-in-hole assembly task can, however, provide valuable insights for developing a strategy to overcome these challenges and perform press-fit tasks. For instance, learning-based approaches, such as reinforcement learning \cite{johannink2019, beltran2020_variable_control, johannsmeier2019, vuong2021}, learning from demonstration \cite{hermann2020, Zhan2020, edward2021} and contact-state recognition \cite{yan2021, Jasim2014} have been successfully applied to learn and perform a peg-in-hole task. Such approaches can be adapted to tackle the press-fit task, as press-fitting can be seen as an instance of the peg-in-hole task. Most of these methods require a large amount of data, however, which can be difficult to collect in real industrial environments and, furthermore, typically do not incorporate mechanisms that enable failure detection and subsequent recovery. This, in turn, limits the applicability of existing approaches to real-world press-fitting scenarios.
    
    \begin{figure}[t]
        \centering
        \includegraphics[width=\linewidth]{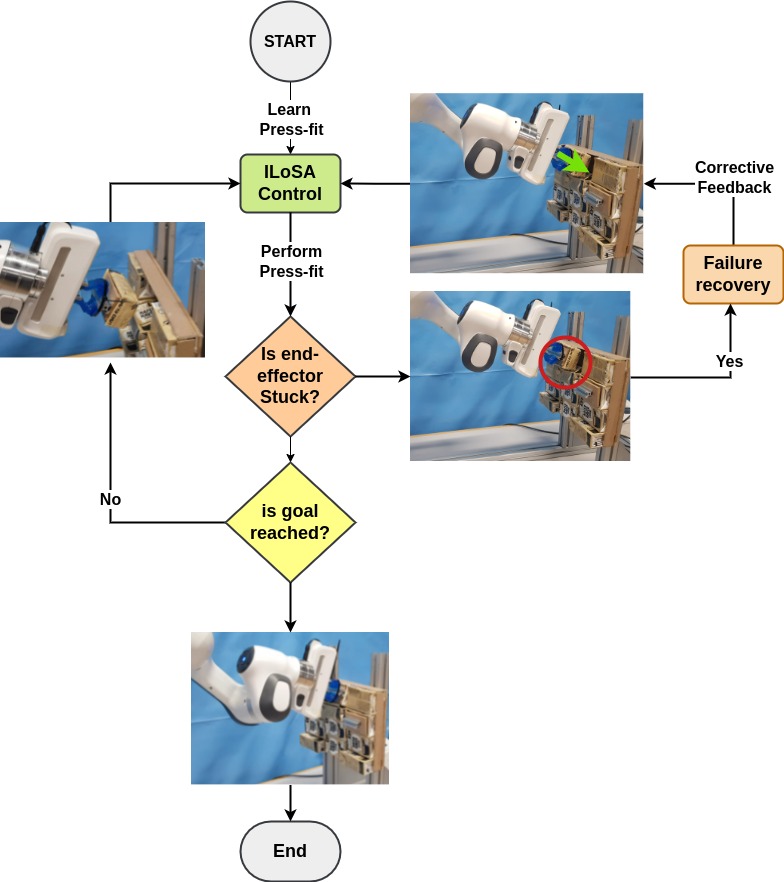}
        \caption{Press-fitting of milk cartons by a mobile manipulator using the proposed Adaptive Compliant Control with Integrated Failure Recovery (ACCIFR). The system comprises three key components: (i) the ILoSA framework for learning and execution (green), (ii) press-fit monitoring via force and distance feedback (yellow), and (iii) a failure recovery mechanism to detect and recover from collisions (orange).}
        \label{fig:process_flow}
    \end{figure}

    This paper presents an approach that we refer to as Adaptive Compliant Control with Integrated Failure Recovery (ACCIFR) to learn and perform a press-fit task using a mobile manipulator. Our approach, illustrated in Fig. \ref{fig:process_flow}, is built upon the ILoSA framework \cite{franzese2021} to learn variable impedance policies from a single demonstration and subsequent user corrections, such that we incorporate a press-fit monitoring system into our approach, which utilizes force and position-based feedback to ensure the reliable press-fitting of milk cartons. Furthermore, our approach extends the ILoSA framework by integrating a failure recovery mechanism that can automatically detect and recover from collisions, inspired by \cite{mitrevski2023_ras}. Concretely, when a collision event occurs, a classifier predicts the contact side using time-series wrench\footnote{Here, wrench data refers to the measurements of forces and torques exerted on a robot's end effector. Such a measurement consists of six components --- three force components and three torque components --- over the $(x, y, z)$-axes.} data, while the recovery mechanism provides corrective feedback based on the predicted collision side, facilitating effective collision recovery.  We demonstrate the generalizability of the proposed approach by performing press-fit tasks using a Franka Emika manipulator and a small container setup, considering variations in the starting position of the end effector, the goal configuration, and the object grasping position. We also evaluate the accuracy of our contact-state recognition classifier in predicting the contact side over varying lengths of time-series history. The results show that ACCIFR improves the performance of the baseline ILoSA, thus suggesting that our approach enables a robot manipulator to learn and perform practical press-fit tasks with the ability to reliably generalize over different environmental variations and recover from collisions using its failure recovery mechanism.

    \section{RELATED WORK}
    \label{sec:related_work}

    The press-fit task is a challenging problem that has received relatively little attention in the context of robotic solutions; however, existing research on the extensively studied peg-in-hole assembly task \cite{suomalainen2022, johannsmeier2019, chernyakhovskaya2020} can provide valuable insights to address this problem. In theory, the press-fit task can be considered as a variation of the peg-in-hole task, where a carton is inserted into a confined space inside a container. In \cite{xu2019}, existing robotics peg-in-hole strategies are classified into two main categories: contact model-based and contact model-free approaches, where the former employ contact-state recognition with compliant control, while the latter use learning through environment interaction or learning from demonstrations.

    Contact-state recognition allows robots to perform manipulation tasks by adjusting their behavior based on the current contact situation. This can be achieved through analytical modeling \cite{whitney1982} or statistical modeling \cite{jakov2012}. Conceptually, analytical modeling for a press-fit task is challenging due to the complexity of modeling numerous constraints, especially the soft body characteristics of the carton for in-contact manipulation, and is generally susceptible to uncertainty. In contrast, statistical modeling directly learns the relationship from gathered samples to predict the contact state without requiring task-specific information. Yan et al. \cite{yan2021} introduce a supervised learning-based contact-state recognition model using support vector machines. This system receives force and torque input from sensors and outputs the corresponding impedance controller and skill parameters to adjust the pose and orientation of the robot's end effector. Jasim and Plapper \cite{Jasim2014} use a Gaussian mixture model and the expectation-maximization algorithm to model input observations (wrench and pose) and estimate the contact state using Bayesian classification. Our proposed approach also uses supervised learning-based contact-state recognition for failure recovery; however, in contrast to \cite{yan2021} and \cite{Jasim2014}, we use a time-series-based classifier to estimate the contact side from a short history of raw wrench data and use this information to generate corrective feedback for failure recovery, making it simpler and more efficient to implement for a press-fit task. 

    Reinforcement learning (RL)-based methods are also widely used to enable robots to learn new tasks and adapt to new situations \cite{rlsutton2018}. Traditional control approaches are typically combined with RL techniques to perform peg-in-hole tasks. Johannink et al. \cite{johannink2019} combine a learnable parametrized policy with a fixed hand-engineered controller, such that the twin delayed deep deterministic policy gradient (TD3) is used as the underlying RL algorithm. Beltran-Hernandez et al. \cite{beltran2020_force_control} use the soft actor-critic (SAC) algorithm to find a policy that generates trajectory commands and the parameters of a force controller based on an estimated goal pose. Learning manipulation skills in combination with RL to perform peg-in-hole tasks is also a common approach in the literature. While some of these frameworks use a fixed set of manipulation primitives for a specific task \cite{johannsmeier2019}, others use a dynamic sequence of manipulation primitives automatically discovered through deep RL \cite{vuong2021}. The sample efficiency, stability, and generalization ability for real-world press-fitting is, however, challenging; most of the above approaches thus suggest using Sim2Real with random exploration and domain randomization \cite{deng2021,beltran2020_variable_control}. However, developing an accurate simulation for press-fit tasks that can represent the contact-rich manipulation and soft-body characteristics of drink cartons is not trivial, while random exploration in the real world can be dangerous.

    Learning from demonstration (LfD) is an alternative promising approach to learning and performing a manipulation task. LfD is often combined with RL to improve data efficiency \cite{abbatematteo2021, cruz2019} due to its ability to bootstrap the agent at the start of the training. Some approaches use only a few demonstrations to learn any manipulation task. For instance, Hermann et al. \cite{hermann2020} propose an adaptive curriculum generation framework that only requires a few expert demonstrations to learn a task in simulation and transfer it to a real robot. Similarly, Zhan et al. \cite{Zhan2020} propose a framework that utilizes only ten expert demonstrations and achieves optimal policies across several manipulation tasks. Recent work in learning from the demonstration is inclined towards one-shot demonstration learning, namely learning any manipulation task through a single demonstration. Johns \cite{edward2021} presents an approach that learns manipulation tasks using a single expert demonstration combined with vision-based self-supervised training from a bottleneck pose (from which the object interaction begins); the approach uses a coarse-to-fine trajectory where the robot approaches the object's bottleneck pose in a coarse manner and then interacts with the object in a fine manner by replaying the end effector velocities recorded during the demonstration. A promising alternative to visual-based imitation learning is presented by Franzese et al. \cite{franzese2021}, which introduces the Interactive Learning of Stiffness and Attractors (ILoSA) framework. ILoSA also utilizes a single demonstration and uses active user corrections to learn diverse manipulation tasks. Here, Gaussian processes are used to learn variable impedance policies, identify uncertainty regions, as well as enable interactive corrections, modulation of stiffness, and active disturbance rejection. Our proposed approach modifies and extends the ILoSA framework to learn and perform the press-fit manipulation task from a single demonstration and user correction, but integrates a monitoring and failure recovery mechanism to perform the press-fit task reliably.

    \section{BACKGROUND: COMPLIANT CONTROL USING ILoSA}
    \label{sec:ilosa}

    The proposed approach is built upon ILoSA \cite{franzese2021}, an interactive imitation learning framework. In this section, we provide a short overview of ILoSA so that our approach, which is described in the next section, can be understood without ambiguities.

    ILoSA uses two main teaching methods: kinesthetic demonstration and teleoperated feedback. We collect a single kinesthetic demonstration $D$ to initialize the policy for the end-effector's intended behavior:
    \begin{gather} \label{eq:d}
         D = \{ (\xi , s) \}
    \end{gather}
    Here, $\xi = \{\phi_0, ..., \phi_T \}$ is a sequence of features $\phi$ defining the state of the robotic system, while $s = \{\Delta x^d, K^d_{s}\}$ represents task-specific information, such that $\Delta x^d$ is the attractor distance recorded in the demonstration and $K^d_{s}$ is the recorded stiffness. The acquired policy can then be executed, such that online corrections $U$ are provided using teleoperated feedback to improve the policy:
    \begin{gather} \label{eq:u}
        U = \{(x_{_{offset}}, y_{_{offset}},  z_{_{offset}})\}
    \end{gather}
    Here, $x_{_{offset}}$ is the directional feedback along the $x$-axis; $y_{_{offset}}$ and $z_{_{offset}}$ are defined equivalently over the $y$- and $z$-axis, respectively.
    
    In ILoSA, a policy is learned using Gaussian processes and can thus represent uncertain regions, which allows interactive corrections and stiffness modulation. Specifically, corrections impact the parameters of the impedance controller by altering the attractor distance, represented by $\Delta x$, and the stiffness of the end effector, represented by $K_s$, as
    \begin{gather}
        \Lambda (q) \ddot{x}  = K_{s} \Delta x - D \dot{x} + f_{ext}
        \label{eq:impedance}
    \end{gather}
    Here, $\Lambda(q)$ is the Cartesian inertia matrix of the physical system, $D$ is the critical damping matrix, and $f_{ext}$ is external force. The hyperparameters of the GPs are determined using a combination of expectation-maximization and the L-BFGS method. Concretely, during the demonstration phase, the hyperparameters are optimized accordingly, but are kept constant during the correction phase, as the correlation is assumed to be invariant.

    ILoSA uses an uncertainty measure to determine if the robot is in an unvisited area, in which case a corrective sample is added to a database. A correction is then distributed among all existing samples that are correlated with the current end effector position using the following update rule, which is used to correct the attractor distance or stiffness (or both) based on interpreting the received feedback:
    \begin{gather}
        \mu(\boldsymbol{x})=\boldsymbol{k}_{*}(\boldsymbol{\xi}, \boldsymbol{x})^{\top}\left(\boldsymbol{K}(\boldsymbol{\xi}, \boldsymbol{\xi})+\sigma_{n}^{2} \boldsymbol{I}\right)^{-1} \boldsymbol{y}=\boldsymbol{A}(\boldsymbol{\xi}, \boldsymbol{x}) \boldsymbol{y} \label{eq:mean}
        \\
        \hspace{-0.09cm}
        \Sigma=k(\boldsymbol{x}, \boldsymbol{x})-\boldsymbol{k}_{*}(\boldsymbol{\xi}, \boldsymbol{x})^{\top}\left(\boldsymbol{K}(\boldsymbol{\xi}, \boldsymbol{\xi})+\sigma_{n}^{2} \boldsymbol{I}\right)^{-1} \boldsymbol{k}_{*}(\boldsymbol{\xi}, \boldsymbol{x}) \label{eq:var}
        \\
        \hspace{-0.1cm}
        \mu+\epsilon_{\mu}=\boldsymbol{A}(\boldsymbol{\xi}, \boldsymbol{x})\left(\boldsymbol{y}+\epsilon_{y}\right) \Rightarrow y_{\text {new }}=\boldsymbol{y}+\boldsymbol{A}(\boldsymbol{\xi}, \boldsymbol{x})^{+} \epsilon_{\mu} \label{eq:update_rule}
    \end{gather}
    Here, $k$ is the variance of $\boldsymbol{x}$, $K$ is the covariance matrix of the training points $\boldsymbol{\xi}$, $\boldsymbol{k}_{*}$ is the covariance between $\boldsymbol{x}$ and the training points, $\sigma_{n}^{2}$ denotes the variance of the Gaussian noise of the training inputs, and $\boldsymbol{y}$ are the training outputs. $K$, $\boldsymbol{k}{*}$, and $k$ are all functions of a kernel, and $\boldsymbol{A}(\boldsymbol{\xi}, \boldsymbol{x})^{+}$ is the pseudoinverse of $\boldsymbol{A}$. The correction provided at $\boldsymbol{x}$ is represented by $\epsilon_{\mu}$. $\boldsymbol{A}^{+}$ acts as a selector that automatically adjusts the modification needed for correlated elements in the database to align with the user's desired corrections.

    ILoSA uses the directional feedback provided by the user to infer changes in the attractor distance or stiffness, which allows for incremental correction of the end effector's dynamics. It concretely uses the teleoperated input feedback to directly determine the attractor distance increment $\Delta x_{\text {inc }}$, while the change in stiffness $K_{s\_inc}$ is obtained as
    \begin{gather} \label{eq:stiffness_update}
        (K_{s}+K_{s\_inc}) \Delta_{\lim }=K_{s}\left|\Delta x+\Delta x_{\text {inc}}\right|
    \end{gather}

    On its own, ILoSA can only be used to learn and execute a policy, but the method does not involve monitoring and recovery mechanisms, which limits its usefulness for practical press-fit applications.
    In this paper, we embed ILoSA within a monitoring and recovery framework that allows reliable execution in the press-fit context.

    \section{COMPLIANT CONTROL WITH FAILURE RECOVERY}
    \label{sec:approach}

    In this work, we propose an approach called Adaptive Compliant Control with Integrated Failure Recovery (ACCIFR) for learning and performing a press-fit task using a mobile manipulator.
    ACCIFR uses ILoSA as a baseline framework to learn a compliant policy for the press-fit task, but incorporates a collision monitoring and recovery mechanism in order to prevent the robot from getting stuck due to collisions with the environment.
    Our approach aims to efficiently execute the press-fit task while maintaining the ability to generalize to task variations. To achieve this, we integrate a failure recovery mechanism that uses contact-state recognition to detect and recover from collisions, namely ACCIFR uses corrective feedback to adjust its policy and recover from a collision. These features enable the ACCIFR algorithm to handle various environmental variations, recover from failures, and lead to successful press-fitting.

    \subsection{Formulation of a Press-Fit Task}
    \label{sec:press_fit}

\begin{figure}[t]
        \centering
        \includegraphics[width=0.30\textwidth]{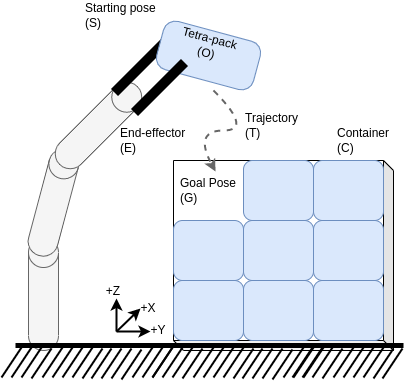}
        \caption{Illustration of a press-fit task}
        \label{fig:press_fit_formulation}
    \end{figure}

    In a container loading process, the press-fit task involves pushing a carton to fit at a designated spot inside the container. Formally, the objective of the press-fit task is to move the robot's end effector $E$ from a starting pose $S$, following a trajectory $T$ in space, to fit an object $O$ at a goal pose $G$, as shown in Fig. \ref{fig:press_fit_formulation}. We assume that $S$ and the estimated goal pose $G$ will be known a priori. Also, we assume that $O$ is grasped in such a way that it can be pushed to fit inside the goal pose. The observable measurements are the end effector pose and the wrench data at each time step. The success of the press-fit task relies on the robot's ability to accurately navigate $T$ and fit $O$ into $G$ in a compliant manner, namely while avoiding collisions with other objects and the walls of the container.

    \subsection{Press-Fit Monitoring and Recovery}
    \label{subsec:goal_monitoring}

    To ensure that the press-fit is achieved correctly, we monitor both the distance between the current end effector pose $\Delta x_{t-1}$ and the goal pose $G$, as well as the force applied along the robot's $x$-direction. The execution is deemed successful when the goal pose is reached within a predefined distance threshold, denoted as $D_{{th}}$, and a desired force threshold, denoted as $F_{{th}}$, is achieved. This helps to ensure that the object is press-fitted securely, which is essential for the successful execution of the task.

\begin{figure}[t]
        \centering
        \includegraphics[width=\linewidth]{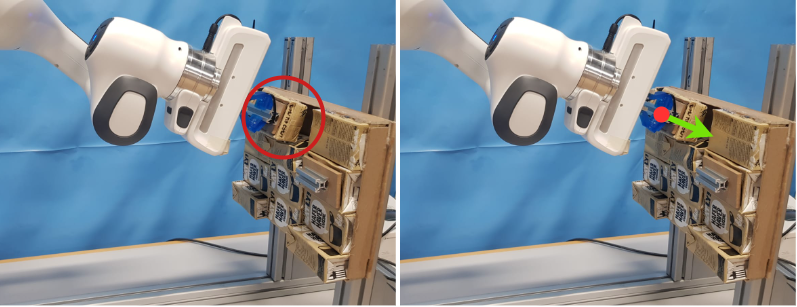}
        \caption{Failure recovery mechanism: (i) a collision is detected (in red), (ii) a classifier predicts the contact side, and (iii) corrective feedback steers the end effector towards the goal (in green) based on the contact-side prediction.}
        \label{fig:failure_recovery}
    \end{figure}

    During execution of the press-fit task, the robot can experience collisions with the environment, particularly with other cartons as well as with the container edges.
    Our failure recovery mechanism, illustrated in Fig. \ref{fig:failure_recovery}, follows a three-step approach to detect and recover from collisions. The first step involves collision detection. Once a collision is detected, a classifier predicts the collision side based on time-series wrench data. Finally, corrective feedback recovers the end effector by driving it towards the goal. We describe each of these steps below.

    \subsubsection{Collision detection}
    \label{sec:collision_detec}

    ILoSA predicts and sets the attractor pose and stiffness at each timestep during active control. We utilize the generated prediction for collision detection, namely we continuously monitor the difference between $\Delta x_{t-1}$ and $\Delta x_{t}$. If there is no difference between the current attractor pose $\Delta x_{t-1}$ and the next predicted attractor pose $\Delta x_{t}$, and the goal pose is not reached, we consider that the end effector has either collided or is stuck.

    \subsubsection{Contact side prediction}
    \label{subsec:classifier_description}

    To predict the side of a collision after it has been detected, we use the \emph{InceptionTime} \cite{ismail2020} deep learning-based time-series classification model.\footnote{We use the implementation of the model provided in the PyTorch Time library for this purpose \url{https://github.com/VincentSch4rf/torchtime}.}
    We employ a deep learning classifier due to its inherent capability to automatically learn complex patterns and representations directly from the raw time-series wrench data, eliminating the need for explicit feature engineering as typically required by classical machine learning approaches. The input to the classifier, denoted as $I_{wrench}$, is multivariate time-series wrench data of the form
    \begin{gather} \label{eq:data_matrix}
        I_{wrench} =
        \begin{bmatrix}
        x_{f_1} & ... & x_{f_n} \\
        y_{f_1} & ... & y_{f_n} \\
        z_{f_1} & ... & z_{f_n} \\
        x_{\tau_1} & ... & x_{\tau_n} \\
        y_{\tau_1} & ... & y_{\tau_n} \\
        z_{\tau_1} & ... & z_{\tau_n}
        \end{bmatrix}
    \end{gather}
    where $(x_f,\ y_f,\ z_f)$ and $(x_\tau,\ y_\tau,\ z_\tau)$ are force and torque readings, respectively, sensed at the end effector, and $n$ represents the length of the wrench data history. The output of the classifier is the predicted contact side label, denoted as $P_{contact}$. For the evaluation in this paper, we collected labeled wrench data for collisions on the left and right sides for simplicity, but the classifier can be extended if other collision categories need to be classified. For training, we collected a wrench data stream for several seconds after the collision occurred. The data was obtained from $80$ trials following a collision and then randomly divided into training and test sets using an $80\% / 20\%$ split ratio. The raw data was preprocessed to ensure it is in a suitable format as in Eq. \ref{eq:data_matrix}; the classifier was trained on the preprocessed data using the cross-entropy loss function and Adam optimizer.

    \subsubsection{Corrective feedback}
    \label{subsec:corrective_feedback}

    In this paper, we use a failure recovery mechanism that is inspired by \cite{mitrevski2023_ras}, where, for parameters that have led to a failure of a parameterized skill, a hypothesis about the failure in terms of violated symbolic relations is found; this hypothesis is then used to guide the subsequent recovery process.
    Concretely, given parameters $\boldsymbol{x}_f$ that have resulted in a failure as well as parameters $\boldsymbol{x}_f'$ that violate relations of a nominal execution model, a corrective set of parameters $\boldsymbol{x}^*$ is identified by moving $\boldsymbol{x}_f$ in a direction away from $\boldsymbol{x}_f'$, as this makes it more likely that the violation of the relations will be remedied.

    For the recovery mechanism we propose in this paper, we use the wrench data classifier described above to ground relations describing collision directions and then use the idea of recovery by moving away from the direction in which a collision is identified.
    Concretely, after a collision is detected, corrective feedback recovers the end effector from the collision by steering it away from the side of the collision and towards the goal pose. We use manually defined corrective feedback instead of varying the feedback magnitude as in \cite{mitrevski2023_ras} since, unlike in \cite{mitrevski2023_ras}, we do not have an underlying model that can evaluate the expected quality of the correction; the feedback magnitude was empirically found for both collision sides. This feedback is then interpreted just as the teleoperated feedback described in Eq. \ref{eq:u}.

    Overall, ACCIFR, summarized in Algorithm \ref{alg:accifr}, is a modified and extended version of ILoSA \cite{franzese2021} that accounts for failure monitoring and recovery, thereby contributing to more reliable press-fit execution.
    \begin{algorithm}[t]
        \caption{Adaptive Compliant Control with Integrated Failure Recovery (ACCIFR). The elements of the ILoSA algorithm are shown in black font; our added components are shown in \textcolor{blue}{blue}.}\label{alg:accifr}
        \begin{algorithmic}[1]
            \Procedure{KinestheticDemonstration}{}
            \While{trajectory recording}
                \State $receive$($x_{t} \rightarrow \xi$)
                \State $\Delta x^{d}(x_{t-1})$ \assign $x_{t} - x_{t-1}$
            \EndWhile
            \EndProcedure
            \State $train(GPs)$
            \Procedure{InteractiveCorrections}{$\xi, \Delta x^{d} , K^{d}_s$}
            \While {not \textcolor{blue}{$success$}}
                \State $receive$($x$)
                \State $[\Delta x, \Sigma]$ \assign GP$_{\Delta x}(x)$
                \State $K_s$ \assign GP$_{K_s}(x)$
                \If{\textcolor{blue}{$feedback$} \label{code:feeback_begin}}
                    \State $[\Delta x_{inc}, K_{s\_{inc}}]$ \assign $interpret$($\textcolor{blue}{feedback}$, $\Delta x$, $K_s$)
                    \If{$\Sigma \geq \Sigma_{Threshold}$}
                        \State $append$($x \rightarrow \xi$, $\Delta x + \Delta x_{inc} \rightarrow \Delta x^d$, $K_s + K_{s\_inc} \rightarrow K^d_s$)
                    \Else
                        \State $correct$($\Delta x_{inc} \rightarrow \Delta x^d, K_{s\_inc} \rightarrow K^d_s$ )\label{code:feedback_end}
                    \EndIf
                    \State $fit(GPs)$
                \EndIf
                \State $\Delta x$ \assign GP$_{\Delta x}(x)$
                \State $K_s$ \assign GP$_{K_s}(x)$
                \If{\textcolor{blue}{$collision$($\Delta x_{t-1}, \Delta x_{t} $)}}
                    \State \textcolor{blue}{$P_{contact}$ \assign $predictContactSide$($I_{wrench}$)} \label{code:predict_contact_side}
                    \State \textcolor{blue}{$feedback$ \assign $correction$($P_{contact}$) \label{code:get_feedback}}
                \Else
                    \State $f_{stable}$ \assign $- \alpha \nabla \Sigma$
                    \State $[\Delta x, K_s]$ \assign $modulation$($\Delta x, K_s , f_{stable} , \Sigma$)
                    \State $send$($\Delta x, K_s$)
                \EndIf
                \State \textcolor{blue}{$success$ \assign $monitor$($\Delta x^{g}$, $\Delta x_{t-1}$, $D_{{th}}$, $F_{{th}}$)} \label{code:monitor_goal}
            \EndWhile
            \EndProcedure
        \end{algorithmic}
    \end{algorithm}

    \section{EVALUATION}
    \label{sec:evaluation}

\begin{figure}[t]
        \centering
        \includegraphics[width=0.40\textwidth]{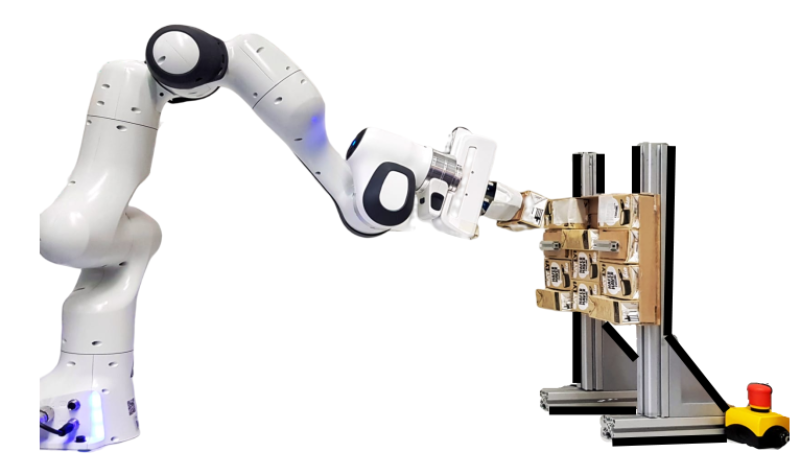}
        \caption{Experimental setup for press-fitting}
        \label{fig:exp_setup}
    \end{figure}

    We conducted experiments to evaluate the performance of our proposed ACCIFR approach compared to the baseline ILoSA approach in terms of its ability to generalize to different variations of the press-fit task. We used a Franka Emika robot manipulator with 7 degrees of freedom, a miniature container setup with milk cartons, and a 3D mouse for user correction, as shown in Fig. \ref{fig:exp_setup}.
    We trained the system using a single demonstration and user correction; the same trained models were used throughout the evaluation. We conducted experiments in three different scenarios of the press-fit manipulation task, considering variations in (i) the robot's starting position, (ii) the goal position, and (iii) the object grasping position. We performed a total of $20$ runs for each scenario with each variation, such that we evaluated the performance using the number of successful runs and the number of collisions encountered during the task.\footnote{The success of each trial was evaluated manually by the experimenter.} In the trials with the robot, we used a contact-state classifier that takes an input history of $290$ wrench measurements, which is about $10s$; however, we also evaluated the prediction performance of the contact-state recognition classifier with varying lengths of wrench data history.

    Our evaluation is based on the following assumptions: (i) before press-fit execution, the robot manipulator starts at a predefined pre-place pose, (ii) an estimated goal pose for placing the carton is given to the robot, (iii) only the arm of the mobile manipulator is moved to perform the press-fit task (fixed base), and (iv) milk cartons have soft-body characteristics.\footnote{A video that illustrates the evaluation process can be found at \url{https://youtu.be/cFChda1Pccc}.}

    \subsection{Generalizability Test I: Starting Position Variation}

    In this scenario, we varied the initial starting position of the end effector in five different ways to fit an object at a fixed goal pose with position $(0.80, -0.05, 0.43)$ and quaternion orientation $(0.58, -0.50, 0.48, -0.40)$ with respect to the robot's base frame. The results in Tab. \ref{tab:variation_in_starting_pos} demonstrate that the ILoSA framework performed exceptionally well in all five cases, namely the end effector successfully reached the goal without collision for each run in each variation. This demonstrates that the baseline ILoSA is capable of dealing with small variations in the starting position, which may suggest that the added recovery by ACCIFR is not needed; however, the benefit of monitoring and recovery becomes clear below.
    \begin{table}[t]
        \centering
        \caption{Performance of ILoSA for variations in the starting position}
        \label{tab:variation_in_starting_pos}
        \resizebox{\linewidth}{!}{%
        \begin{tabular}{|c|l|c|}
        \hline
        \textbf{S.No.}& \textbf{Variation} & \textbf{\begin{tabular}[c]{@{}l@{}}Goal reached (out of 20)\end{tabular}}\\ \hline
        1 & \begin{tabular}[c]{@{}l@{}}position: $(0.74, -0.05, 0.43)$\\          orientation: $(0.58, -0.50, 0.48, -0.40)$\end{tabular} & 20\\ \hline
        2 & \begin{tabular}[c]{@{}l@{}}position: $(0.74, -0.05, \mathbf{0.52})$\\ orientation: $(0.58, -0.50, 0.48, -0.40)$\end{tabular} & 20\\ \hline
        3 & \begin{tabular}[c]{@{}l@{}}position: $(0.74, \mathbf{0.01}, 0.43)$\\  orientation: $(0.58, -0.50, 0.48, -0.40)$\end{tabular} & 20\\ \hline
        4 & \begin{tabular}[c]{@{}l@{}}position: $(0.74, \mathbf{-0.14}, 0.43)$\\ orientation: $(0.58, -0.50, 0.48, -0.40)$\end{tabular} & 20\\ \hline
        5 & \begin{tabular}[c]{@{}l@{}}position: $(\mathbf{0.59, -0.14, 0.52})$\\ orientation: $(0.58, -0.50, 0.48, -0.40)$\end{tabular} & 20\\ \hline
        \end{tabular}%
        }
    \end{table}

    \subsection{Generalizability Test II: Goal Configuration Variation}

\begin{figure}[t]
        \centering
        \includegraphics[width=\linewidth]{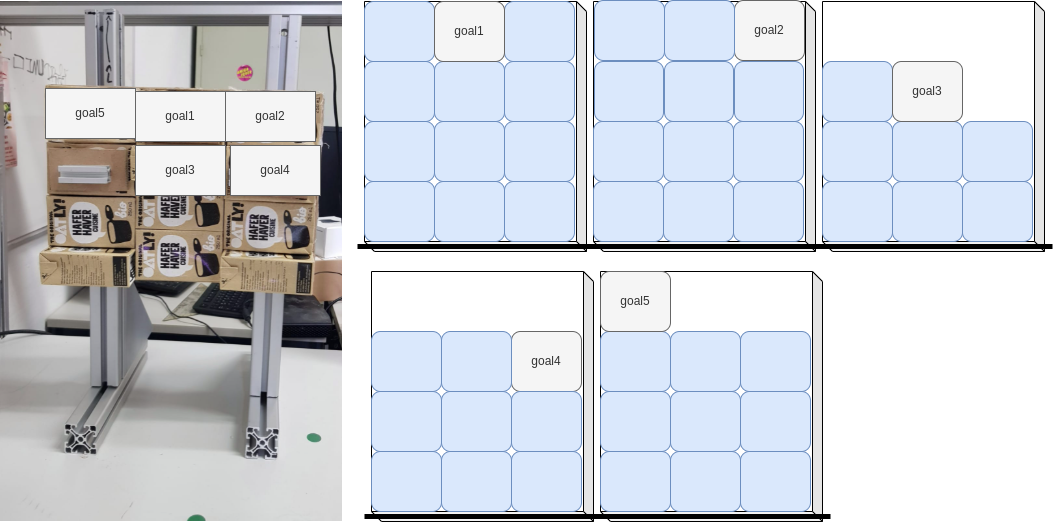}
        \caption{Validation scenario evaluating the generalizability of ACCIFR across different goal configurations, representing contact situations from various directions.}
        \label{fig:variation_goal_config}
    \end{figure}

    In this scenario, we compare the performance of ACCIFR with the baseline ILoSA to perform press-fitting with five different goal configurations, as shown in Fig. \ref{fig:variation_goal_config}. For this, we modified the algorithms to predict the next Cartesian pose of the end effector in the \emph{local} frame to reach the goal pose, which means that the end effector has to only move by a certain distance (in the $x$-direction) to reach the goal. 
    \begin{table}[t]
        \centering
        \caption{Comparision of ILoSA and ACCIFR for variations in the goal configuration}
        \label{tab:goal_config_comp}
        \begin{tabular}{|c|c|cc|}
        \hline
        \multicolumn{1}{|c|}{\multirow{2}{*}{\textbf{S.No.}}} & \multicolumn{1}{c|}{\multirow{2}{*}{\textbf{Variation}}} & \multicolumn{2}{c|}{\textbf{Goal reached (out of 20)}} \\ \cline{3-4} 
        \multicolumn{1}{|c|}{} & \multicolumn{1}{c|}{} & \multicolumn{1}{c|}{\textbf{ILoSA}} & \textbf{ACCIFR} \\ \hline
        1 & $goal1$ & \multicolumn{1}{c|}{20} & 20 \\ \hline
        2 & $goal2$ & \multicolumn{1}{c|}{20} & 20 \\ \hline
        3 & $goal3$ & \multicolumn{1}{c|}{20} & 20 \\ \hline
        4 & $goal4$ & \multicolumn{1}{c|}{0} & 14 \\ \hline
        5 & $goal5$ & \multicolumn{1}{c|}{0} & 20 \\ \hline
        \end{tabular}%
    \end{table}

    As shown in Tab. \ref{tab:goal_config_comp}, the proposed ACCIFR approach outperformed ILoSA regarding successful runs. ILoSA performed well for the default goal pose ($goal1$) but struggled to maintain stability while attempting to reach the other goal poses ($goal2$ and $goal3$), namely it did not reach some goal poses due to collisions. On the other hand, using ACCIFR, the end effector could recover from collisions and reach the goal pose in most cases. Nevertheless, the end effector could not reach $goal4$ in $6$ out of $20$ runs. We hypothesize that this was caused by the impedance control parameters learned from the user corrections, which was supported by the observation that adding new user corrections improved the performance. The results of this experiment demonstrate that the recovery mechanism effectively enabled the end effector to reach different goal poses, but also that appropriate recording of user corrections is essential for improving the system's performance.

    \subsection{Generalizability Test III: Variation in the Object Grasping Position}

\begin{figure}[t]
        \centering
        \includegraphics[width=0.9\linewidth]{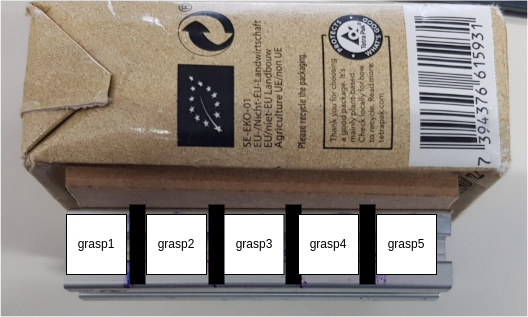}
        \caption{Validation scenario evaluating the generalizability of ACCIFR across different object grasping positions, representing pose uncertainty during grasping.}
        \label{fig:var_obj_grasp}
    \end{figure}

    While performing the press-fit task, the carton might be grasped at a different grasping position than the demonstration with which it is trained, which can also lead to unforeseen collisions with the environment. In this experiment, we compare the performance of ACCIFR with the baseline ILoSA to perform a press-fit task with five different grasping position variations, as shown in Fig. \ref{fig:var_obj_grasp}.

    The experimental results in Tab. \ref{tab:obj_grasp_comp} demonstrate that the proposed ACCIFR approach significantly outperforms the baseline ILoSA regarding generalization to different grasping positions.
    \begin{table}[t]
        \centering
        \caption{Comparision of ILoSA and ACCIFR for variations in the object grasping position}
        \label{tab:obj_grasp_comp}
        \begin{tabular}{|c|c|cc|}
        \hline
        \multicolumn{1}{|c|}{\multirow{2}{*}{\textbf{S.No.}}} & \multicolumn{1}{c|}{\multirow{2}{*}{\textbf{Variation}}} & \multicolumn{2}{c|}{\textbf{Goal reached (out of 20)}} \\ \cline{3-4} 
        \multicolumn{1}{|c|}{} & \multicolumn{1}{c|}{} & \multicolumn{1}{c|}{\textbf{ILoSA}} & \textbf{ACCIFR} \\ \hline
        1 & $grasp1$ & \multicolumn{1}{c|}{0} & 19 \\ \hline
        2 & $grasp2$ & \multicolumn{1}{c|}{0} & 20 \\ \hline
        3 & $grasp3$ & \multicolumn{1}{c|}{0} & 20 \\ \hline
        4 & $grasp4$ & \multicolumn{1}{c|}{20} & 20 \\ \hline
        5 & $grasp5$ & \multicolumn{1}{c|}{0} & 20 \\ \hline
        \end{tabular}%
    \end{table}
    In particular, ILoSA reached the goal pose without collision only when the object was held at the same grasping position as in the demonstration, but it failed in all other variations. In contrast, the ACCIFR approach was successful in all five grasping position variations and recovered from collisions in all cases due to the failure recovery mechanism. The only failure for ACCIFR occurred in the case of $grasp1$, where the grasped object slipped from the gripper in one trial, thus preventing the end effector from reaching the goal pose. These results demonstrate the enhanced adaptability, stability, and collision recovery capabilities of the proposed ACCIFR approach for the press-fit task with variations in the grasping position.

    \subsection{Contact-State Classifier Analysis}

    To examine the effect of the length of the wrench data history on the contact-state classifier and determine the optimal length of wrench data, we also trained multiple versions of the classifier with different lengths of wrench data history.
    Based on the results shown in Tab. \ref{tab:time_length_data}, it can be seen that the contact-state classifier used in ACCIFR can accurately predict the side of a collision regardless of the length of the wrench data history used. In particular, the classification accuracy remains at $100\%$ for all time windows tested, indicating that the classifier can correctly predict the contact side even when using wrench data with a short history length. This could be attributed to the classifier's ability to extract highly discriminative features from the wrench data, facilitated by the \emph{InceptionTime} model; this allows for the prediction of the contact side to be generated quickly, which enables the robot to swiftly recover from failures, thereby minimizing the impact of collisions.
    The high accuracy can also be explained by the structure of the experimental setup, where collision profiles remain static between trials; however, as this is a fairly accurate model of a real container loading scenario, the results suggest that the contact-state classifier can also be useful for the real-world press-fitting task.
    \begin{table}[t]
        \centering
        \caption{Influence of the length of wrench data history on the contact-state classification accuracy}
        \label{tab:time_length_data}
        \resizebox{\linewidth}{!}{%
        \begin{tabular}{|c|c|c|c|}
        \hline
        \textbf{S. No.} & \textbf{\begin{tabular}[c]{@{}c@{}}Time \\ (in sec)\end{tabular}} & \textbf{\begin{tabular}[c]{@{}c@{}}Average length of \\ wrench data history\end{tabular}} & \textbf{\begin{tabular}[c]{@{}c@{}}Classification \\ accuracy (in \%)\end{tabular}} \\ \hline
        1 & 10 & 290 & 100 \\ \hline
        2 & 5 & 147 & 100 \\ \hline
        3 & 2 & 59 & 100 \\ \hline
        4 & 1 & 29 & 100 \\ \hline
        5 & 0.5 & 15 & 100 \\ \hline
        6 & 0.2 & 6 & 100 \\ \hline
        7 & 0.1 & 3 & 100 \\ \hline
        8 & 0.05 & 1 & 100 \\ \hline
        \end{tabular}%
        }
    \end{table}

    \section{DISCUSSION AND CONCLUSIONS}
    \label{sec:conclusion}

    The proposed approach, which we refer to as Adaptive Compliant Control with Integrated Failure Recovery (ACCIFR), enables a mobile manipulator to perform a press-fit task. The approach learns the task using a single demonstration and user corrections through the ILoSA framework, while the failure recovery mechanism enables the end effector to avoid getting stuck after colliding with objects, ultimately steering it towards the goal location. Regardless of the initial expert demonstration, ACCIFR's ability to generalize to different variations can be attributed to both ILoSA and the integrated failure recovery mechanism. Moreover, the experimental evaluation indicated that ACCIFR outperformed ILoSA in terms of generalization performance. Concretely, ACCIFR achieved a success rate of $90\%$ for variations in the goal configuration, whereas ILoSA achieved only $60\%$. This is also visible in the case of variations in the object grasping position, where ACCIFR achieved an almost perfect success rate, while ILoSA only achieved a success rate of $20\%$. Our supervised learning-based contact-state classifier exhibited good performance across varying lengths of time-series data, ranging from an average length of $290$ to $1$, indicating its ability to capture the necessary information for predicting the contact side.

    The work presented in this paper focused solely on automating the press-fit task and thus assumes that the mobile manipulator will always start from a pre-place pose, that it knows the estimated goal pose, and that it only uses the arm during the press-fit execution. These assumptions were made due to the limited scope of our study, as the press-fit task is just one aspect of a larger research project.\footnote{The project MASON aims to automate the shipping container loading process using a custom mobile manipulator.} Additional components for automating the shipping container loading process with a custom mobile manipulator are under current development.
    
    Our evaluation of the proposed approach revealed some limitations as well. The controller showed erratic behavior during some trials, which may affect the robot's reliability in performing a press-fit task; future work should thus focus on improving and optimizing the controller for enhanced stability and accuracy. We also observed a hardware limitation in the system, where the object slipped from the end effector during the press-fit task. Using a gripper that can reliably hold the object in place (potentially custom-designed) is essential, especially when performing in-contact manipulation with drink cartons. Another limitation is that our failure recovery mechanism can only recover from collisions on the left or right side; future work should generalize failure recovery behaviors beyond left and right collisions, for instance by incorporating self-exploration and active learning, which could enable the robot to learn from its experiences and actively seek new knowledge to improve its performance. Finally, future work could compare the performance of our time-series classifier with classical learning approaches to assess the practical need for a deep learning-based approach.

\addtolength{\textheight}{-11cm}   


\section*{ACKNOWLEDGMENT}

We would like to thank Vincent Scharf and Konstantin Z{\"a}hl for their support in the experiments.


\bibliographystyle{IEEEtran}
\bibliography{references}

\end{document}